\documentclass[conference]{IEEEtran}
\IEEEoverridecommandlockouts
\usepackage{cite}
\usepackage{amsmath,amssymb,amsfonts}
\usepackage{algorithmic}
\usepackage{graphicx}
\usepackage{textcomp}
\usepackage{booktabs}
\usepackage{xcolor}
\usepackage{multirow}
\usepackage{caption} 
\usepackage{subcaption}
\usepackage{hyperref}

\def\BibTeX{{\rm B\kern-.05em{\sc i\kern-.025em b}\kern-.08em
    T\kern-.1667em\lower.7ex\hbox{E}\kern-.125emX}}
\begin{document}

\title{Extract Information from Hybrid Long Documents Leveraging LLMs: A Framework and Dataset}

\author{
\IEEEauthorblockN{Chongjian Yue\textsuperscript{2, *}, Xinrun Xu\textsuperscript{3, *}, Xiaojun Ma\textsuperscript{1, \dag}, Lun Du\textsuperscript{4, \dag}
\\
Zhiming Ding\textsuperscript{3}, Shi Han\textsuperscript{1}, Dongmei Zhang\textsuperscript{1}, Qi Zhang\textsuperscript{1}}
\IEEEauthorblockA{
\textsuperscript{1}{Microsoft}, Beijing, China
\textsuperscript{2}{School of Software \& Microelectronics, Peking University}, Beijing, China\\
\textsuperscript{3}{Institute of Software, Chinese Academy of Sciences}, Beijing, China
\textsuperscript{4}{Ant Group}, Beijing, China
}
Email:
2201210695@stu.pku.edu.cn, xuxinrun20@mails.ucas.ac.cn, xiaojunma@microsoft.com, dulun2834@gmail.com
\\
zhiming@iscas.ac.cn, \{shihan, dongmeiz, zhang.qi\}@microsoft.com
% zhiming@iscas.ac.cn
% shihan@microsoft.com
% dongmeiz@microsoft.com
% zhang.qi@microsoft.com
}
\maketitle
\renewcommand{\thefootnote}{\fnsymbol{footnote}}
\footnotetext[1]{Co-1st Authors and work done as interns at MSRA.} 
\footnotetext[2]{Corresponding Authors.}
\renewcommand{\thefootnote}{\arabic{footnote}}

\maketitle

\begin{abstract}
Large Language Models (LLMs) demonstrate exceptional performance in textual understanding and tabular reasoning tasks.
However, their ability to comprehend and analyze hybrid text, containing textual and tabular data, remains unexplored.
The hybrid text often appears in the form of hybrid long documents (HLDs), which far exceed the token limit of LLMs. 
Consequently, we apply an \textbf{A}utomated \textbf{I}nformation \textbf{E}xtraction framework (\textbf{AIE}) to enable LLMs to process the HLDs and carry out experiments to analyse four important aspects of information extraction from HLDs.
Given the findings:
1) The effective way to select and summarize the useful part of a HLD.
2) An easy table serialization way is enough for LLMs to understand tables.
3) The naive AIE has adaptability in many complex scenarios.
4) The useful prompt engineering to enhance LLMs on HLDs.
To address the issue of dataset scarcity in HLDs and support future work, we also propose the \textbf{Fi}nancial Reports \textbf{N}umerical \textbf{E}xtraction (\textbf{FINE}) dataset.
The dataset and code are publicly available in the attachments.
\end{abstract}

\begin{IEEEkeywords}
Information Extraction (IE), Large Language Models (LLMs), Hybrid Long Documents, Financial Reports
\end{IEEEkeywords}

\section{Introduction}

Large Language Models (LLMs) have demonstrated exceptional capabilities in understanding, analyzing and reasoning textual and tabular data independently, as evidenced by studies like \cite{wei2023chainofthought,he2024text2analysis,xu2024surveygameplayingagents,wu2024strago}. However, their application to hybrid long documents (HLDs), which intricately weave together textual and tabular content, remains relatively unexplored \cite{yue2023enabling}. This work addresses this gap by investigating the potential of LLMs for information extraction from HLDs, introducing an Automated Information Extraction framework called AIE.

Given the length constraints of LLMs, directly processing entire HLDs is impractical and simple truncation leads to significant information loss \cite{kojima2023large,chen2023large,ye2023large,tan2024cradleempoweringfoundationagents}. AIE tackles this challenge by splitting HLDs into manageable segments and leveraging LLMs to extract relevant information from these segments. Our research delves into four key challenges associated with HLD information extraction:
\textbf{C1. Effective Selection and Summarization of Relevant Segments:} With keyword-related information scattered across segments, effectively identifying and summarizing relevant content is crucial. We compare the ``Refine" and ``Map-Reduce" summarization strategies, exploring the trade-off between accuracy and efficiency. We also investigate the impact of varying the number of retrieved segments based on keyword similarity.
\textbf{C2. Optimal Table Serialization for LLMs:} Tables are integral to HLDs, but LLMs can't directly interpret tabular data. We explore different table serialization formats to identify the optimal representation for LLM comprehension. Our findings indicate that a simple, less hierarchical format is sufficient for LLMs to effectively understand tabular information.
\textbf{C3. Adaptability of AIE: } HLDs encompass a diverse range of domains and complexities. We evaluate AIE's adaptability through experiments assessing its performance across various domains, its handling of ambiguous expressions, and its compatibility with LLMs possessing different capabilities.
\textbf{C4. Prompt Engineering for Enhanced Information Extraction: } Prompt engineering significantly influences LLM performance. We investigate and present effective prompt engineering techniques tailored for AIE in HLD information extraction. These include strategies for numerical precision enhancement, keyword completion leveraging document metadata, and effective few-shot learning approaches.

The subsequent sections detail the AIE framework, incorporating globally optimal settings determined through our analysis (Section \ref{sec:framework}). We introduce the experimental dataset and evaluation metrics (Section \ref{sec:dataset}) and present a comprehensive performance evaluation of AIE. Finally, we delve into a detailed analysis of the impact of various design choices on AIE's effectiveness.

\section{AIE Framework} \label{sec:framework}

\begin{figure*}[h]
\centerline{\includegraphics[width=0.685\linewidth]{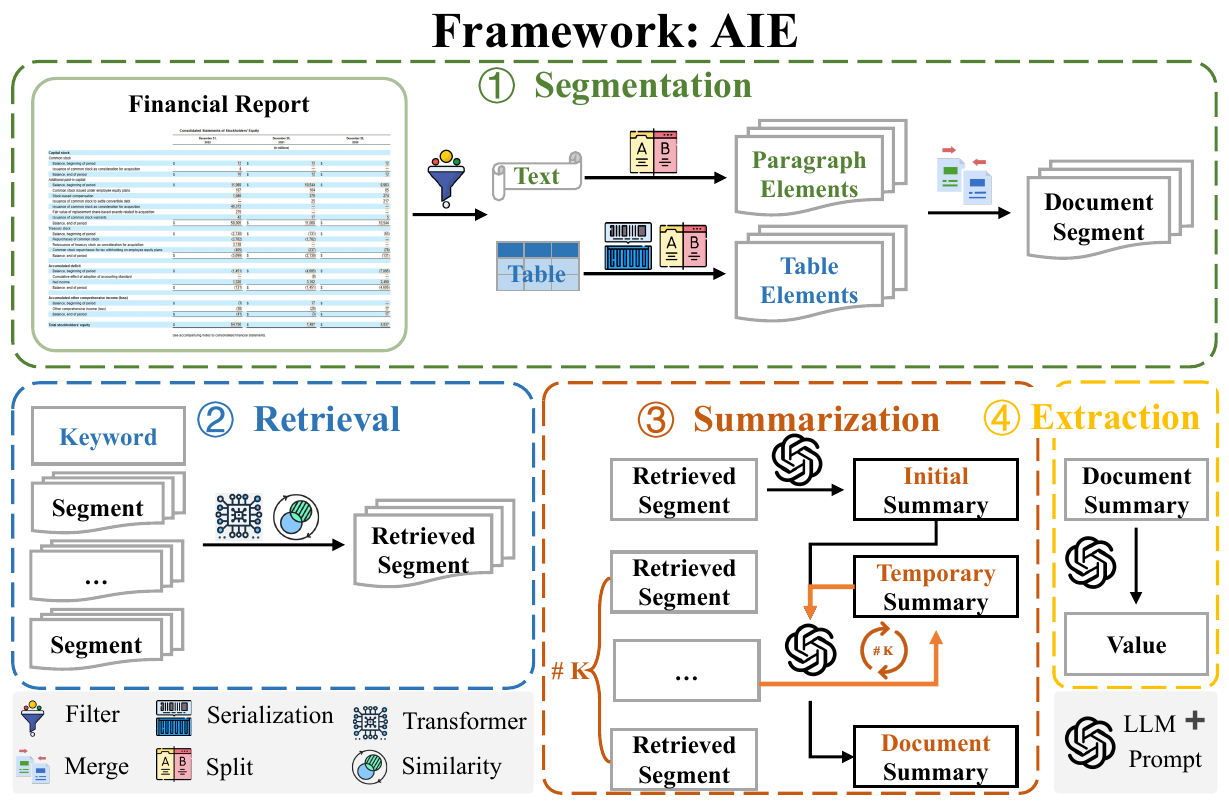}}
% \caption{The AIE framework illustrates the end-to-end IE process, consisting of four modules: \textbf{Segmentation}, dividing lengthy documents into short segments; \textbf{Retrieval}, selecting the most relevant segments related to the given keyword; \textbf{Summarization}, using LLMs to generate a concise summary of relevant information; and \textbf{Extraction}, extracting the keyword-corresponding value from the summary. This framework is exemplified using financial reports.}
\caption{The AIE framework illustrates the end-to-end IE process, consisting of four modules: Segmentation, Retrieval, Summarization, and Extraction, extracting the keyword-corresponding value from the summary.}
\label{fig:framework}
\end{figure*}

To enable LLMs to process HLDs, we propose the Automated Information Extraction (AIE) framework, consisting of four modules: Segmentation, Retrieval, Summarization, and Extraction (Figure~\ref{fig:framework}). AIE segments documents, retrieves keyword-relevant segments based on embedding similarity, summarizes these segments and finally extracts the target value.

The \textbf{Segmentation} \textbf{Module} splits HLDs into manageable segments for LLMs. It involves \textbf{Serialization}, converting tables into text using the simple yet effective \textit{PLAIN} method; \textbf{Split}, dividing overlong elements into smaller sub-elements to avoid information loss; and \textbf{Merge}, concatenating adjacent small elements to maintain semantic relationships.

The \textbf{Retrieval} \textbf{Module} employs embedding-based retrieval \cite{li2021embedding} to avoid processing all segments \cite{wang2024can,yue2024mllm}. It calculates the similarity between each segment and the keyword using the Sentence-Transformer model \cite{reimers2019sentence} and retrieves the top-ranked segments.

The \textbf{Summarization} \textbf{Module} uses LLMs to generate a concise summary from the retrieved segments, capturing relevant information related to the keyword. We employ the \textbf{Refine Strategy}, which iteratively updates an evolving summary with information from each segment.

Finally, the \textbf{Extraction} \textbf{Module} extracts the precise numerical value from the generated summary using LLMs and a tailored \textit{Extraction Prompt}. 

We utilize three key prompt engineering techniques to enhance AIE's performance: \textbf{Numerical Precision Enhancement} ensures accurate numerical extraction, crucial for financial analysis, using Direct and Shot-Precision methods; \textbf{Keyword Completion} improves IE accuracy by completing incomplete user-provided keywords using document metadata; and \textbf{Few-Shot Learning} guides LLMs to understand the task effectively through a single, well-designed shot.

\section{Dataset and Evaluation Metrics}
\label{sec:dataset}

\begin{table}[h]  
    \centering  
    % \small  
     \caption{Basic statistics for datasets.} 
    \begin{tabular}{l|ccc}  
        \hline  
        \textbf{Dataset} & \textbf{FINE} & \textbf{WIKIR} & \textbf{MPP} \\ \hline   
          
        Max \# tokens & 234,900 & 58,512 & 123,105\\   
          
        Min \# tokens & 13,022 & 13,548 & 3,672\\   
          
        Avg. \# tokens & 59,464.3 & 30,922.1 & 17,553.05 \\ \hline  
    \end{tabular}  
    \label{tab:statistics}
\end{table}

To evaluate LLMs' capacity for HLD comprehension, we conduct experiments using three datasets:
\textbf{FINE}: A new dataset with financial KPIs extracted from SEC's EDGAR (Table~\ref{tab:statistics}).
\textbf{WIKIR}~\cite{DBLP:journals/corr/abs-2011-03228}: Extracts key-value pairs from Wikipedia pages and Wikidata.
\textbf{MPP}~\cite{polak2023flexible}: Extracts chemical material properties from scientific papers.

\textbf{FINE} utilizes the Relative Error Tolerance Accuracy (RETA) metric due to varying numerical precision in financial reports. RETA considers predictions correct if their relative error is within a specified threshold (e.g., RETA X\% means predictions with a relative error of no more than X\% are considered correct)
By setting different RETA levels, we can assess the model's performance according to various practical requirements and gain a comprehensive understanding of its capabilities in IE from financial reports.
\textbf{WIKIR} and \textbf{MPP} use Accuracy (Acc) as their ground truth values don't exhibit precision variations.

\section{Experiment}

\begin{figure}[h]
    \centering
    \begin{subfigure}{0.49\columnwidth}
        \centering
        \includegraphics[width=\textwidth]{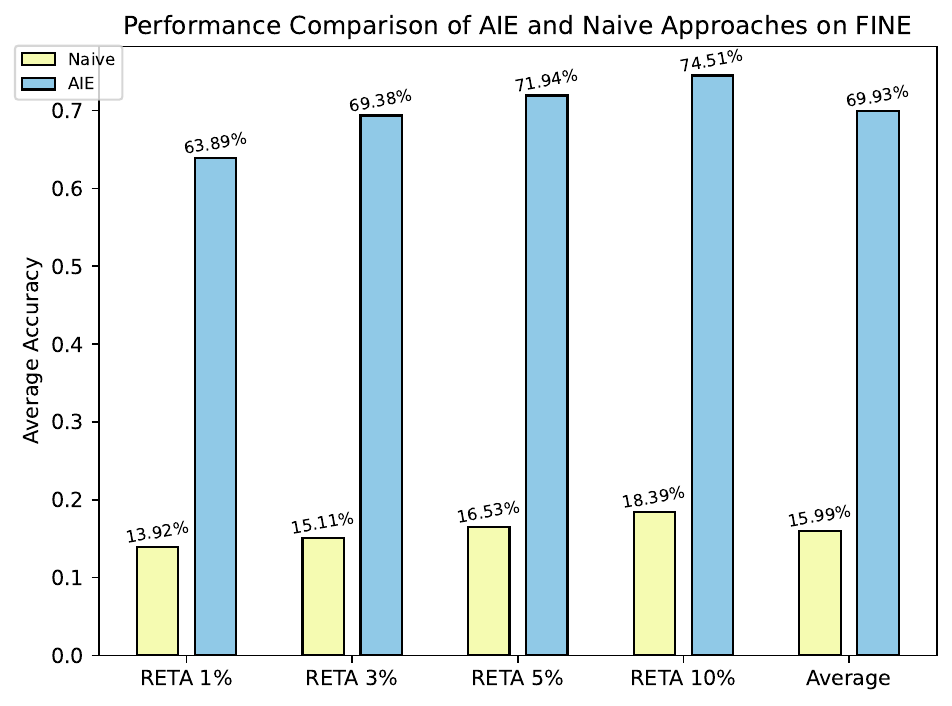}
        \caption{Comparison using GPT-3.5.}
        \label{fig: exp_main_result}
    \end{subfigure}
    \hfill
    \begin{subfigure}{0.49\columnwidth}
        \centering
        \includegraphics[width=\textwidth]{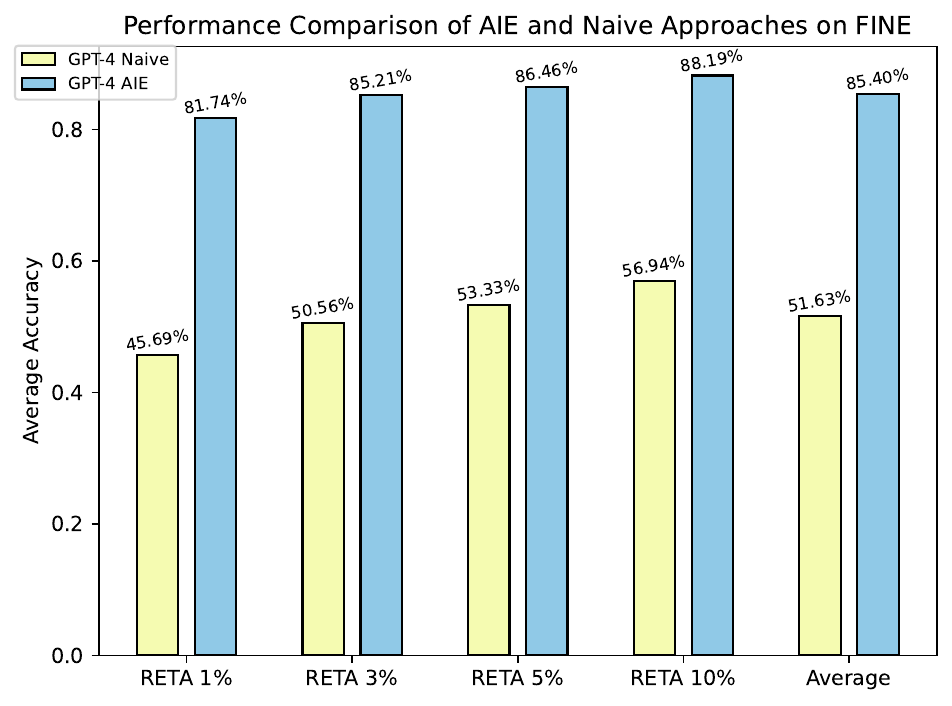}
        \caption{Comparison using GPT-4.}
        \label{fig: result_gpt4}
    \end{subfigure}
    \caption{Comparison of the Naive method and AIE at different RETA levels on FINE.}
\end{figure}

\begin{figure}[h]
\centering
\includegraphics[width=0.6\columnwidth]{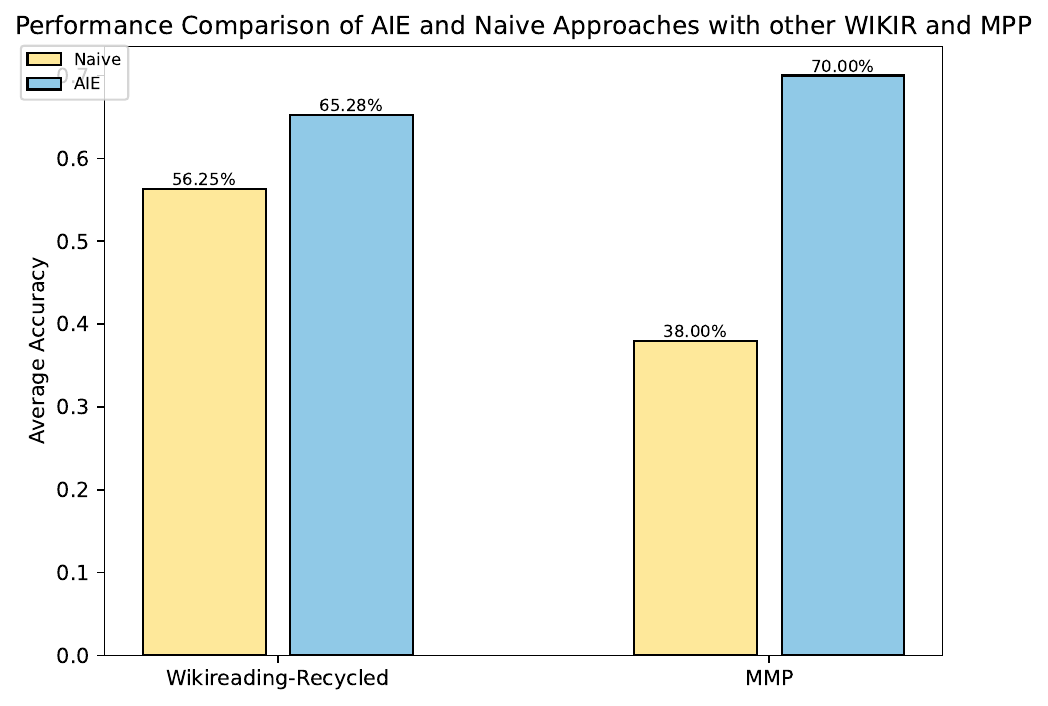}
\caption{Comparison of the Naive method and AIE on WIKIR and MPP using GPT-3.5.}
\label{fig: new_dataset_result}
\end{figure}

We compare our proposed Automated Information Extraction (AIE) framework with a naive LLM-based approach across all three datasets (Figure ~\ref{fig: exp_main_result}, Figure ~\ref{fig: new_dataset_result}). 
AIE consistently outperforms the naive method, demonstrating its effectiveness in extracting information from diverse HLDs. Notably, AIE's accuracy advantage widens under stricter RETA levels on FINE, highlighting its ability to achieve higher precision.

To assess AIE's adaptability across LLMs, we evaluated it using GPT-4, the current leading LLM. Focusing on the FINE dataset, Figure \ref{fig: result_gpt4} shows that AIE consistently outperforms the naive approach across all RETA levels, demonstrating its adaptability and robustness even with more powerful LLMs.

To evaluate AIE's ability to handle ambiguous concepts common in HLDs, we compared its performance to the naive method on two keyword sets: (\textit{Revenue} vs. \textit{Total Net Sales}) and (\textit{Total Equity} vs. \textit{Total Stockholders' Equity}). We used the Relative Percentage Difference (RPD) in average accuracy between the methods across RETA levels:

$$RPD_{X-Y} = \frac{abs(Acc_X - Acc_Y)}{average(Acc_X, Acc_Y)}$$

\begin{figure*}[h]
\centering
\includegraphics[width=0.7\linewidth]{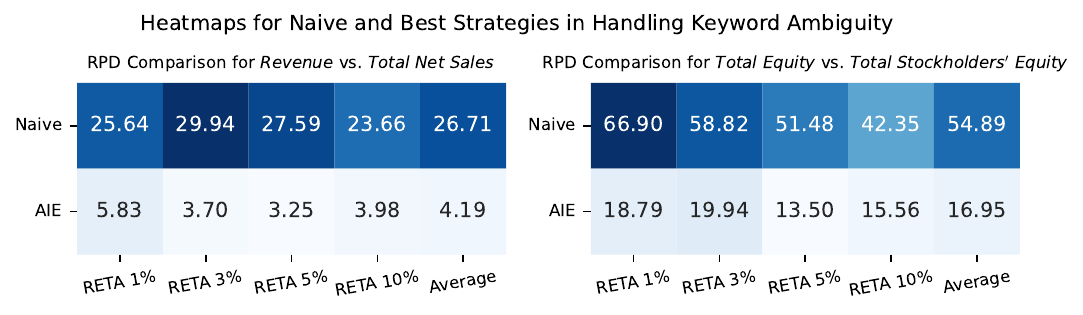}
\caption{Exploring the Capability to Handle Keyword Ambiguity: Comparison of Naive and AIE on RPD}
\label{fig:heatmap_keyword_ambiguity}
\end{figure*}

As shown in Figure \ref{fig:heatmap_keyword_ambiguity}, AIE consistently demonstrates superior performance in handling keyword ambiguity across all RETA levels. AIE achieves a 22.52\% lower average RPD for the ``Revenue" pair and a 37.94\% lower average RPD for the ``Total Equity" pair compared to the naive method. This highlights AIE's effectiveness in disambiguating concepts within HLDs.

\begin{table}[ht]        
\centering        
\caption{Accuracy among table serialization formats.}
\fontsize{8}{10}\selectfont 
\setlength{\tabcolsep}{2pt}
\begin{tabular}{l|cccc|c}        
\hline        
& \textbf{RETA 1\%} & \textbf{RETA 3\%} & \textbf{RETA 5\%} & \textbf{RETA 10\%} & \textbf{Average} \\ \hline    
\textbf{PLAIN} & \textbf{0.6389} & \textbf{0.6938} & \textbf{0.7194} & \textbf{0.7451} & \textbf{0.6993} \\        
\textbf{CSV} & 0.6264 & 0.6889 & 0.7132 & 0.7361 & 0.6911 \\        
\textbf{XML} & 0.3951 & 0.4507 & 0.4729 & 0.5069 & 0.4564 \\       
\textbf{HTM}L & 0.4542 & 0.5000 & 0.5208 & 0.5590 & 0.5085 \\ \hline        
\end{tabular} 
\label{tab:serialization_formats}
\end{table}

\begin{table}[ht]      
\centering  
% \caption{Accuracy for different retrieval quantities (R@n).} 
\caption{Accuracy comparison for different retrieval quantities (R@n) across various RETA levels.} 
\fontsize{8}{10}\selectfont 
\setlength{\tabcolsep}{2pt} 
\begin{tabular}{l|cccc|c}      
\hline      
 & \textbf{RETA  1\%} & \textbf{RETA  3\%} & \textbf{RETA  5\%} & \textbf{RETA  10\%} & \textbf{Average} \\ \hline      
\textbf{R@1} & 0.4757 & 0.5278 & 0.5444 & 0.5694 & 0.5293 \\ 
\textbf{R@2} & 0.6188 & 0.6736 & 0.6931 & 0.7118 & 0.6743 \\ 
\textbf{R@3} & \textbf{0.6389} & \textbf{0.6938} & \textbf{0.7194} & \textbf{0.7451} & \textbf{0.6993} \\      
\textbf{R@5} & 0.6160 & 0.6799 & 0.7062 & 0.7306 & 0.6832 \\     
\textbf{R@7} & 0.5917 & 0.6521 & 0.6722 & 0.7090 & 0.6563 \\       
\textbf{No R} & 0.3757 & 0.4986 & 0.5201 & 0.5514 & 0.4865 \\ \hline      
\end{tabular}      
\label{tab:retrieval_quantity}
\end{table} 

To enable LLMs to process tables, we evaluate four serialization methods: PLAIN, CSV, XML, and HTML. 
While XML and HTML retain hierarchical table structure using tags, they increase token count and potential table fragmentation. Table \ref{tab:serialization_formats} shows that PLAIN and CSV, which prioritize conciseness, outperform XML and HTML in accuracy. This suggests that preserving complete semantic information without excessive structural details is crucial for effective LLM-based table understanding.

We analyze the impact of retrieved segment quantity (R@n) on accuracy (Table \ref{tab:retrieval_quantity}). R@3 achieves the highest accuracy across all RETA levels. While retrieving more segments initially improves accuracy (R@1 to R@3), exceeding this threshold leads to a decline, likely due to the introduction of noise or irrelevant information.

\begin{figure}[h]
\centerline{\includegraphics[width=0.5\linewidth]{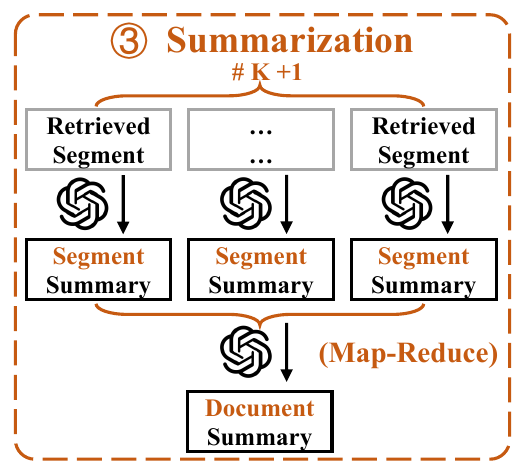}}
% \caption{Illustration of the Map-Reduce Strategy, comprising two stages: Map, generating individual segment summaries, and Reduce, combining these summaries to form a single document summary.}
\caption{Illustration of the Map-Reduce Strategy.}
\label{fig:map-reduce}
\end{figure}

\begin{table*}[htbp]      
\centering     
\caption{Accuracy comparison between Map-Reduce and Refine strategies across various RETA levels.}   
\small
\begin{tabular}{l|cccc|cc}      
\hline      
& \textbf{RETA 1\%} & \textbf{RETA 3\%} & \textbf{RETA 5\%} & \textbf{RETA 10\%} & \textbf{Average} & \textbf{Time (s\textbackslash sample)} \\ \hline      
\textbf{Map-Reduce} & 0.5375 & 0.5729 & 0.5958 & 0.6299 & 0.5840 &  \textbf{13.34} \\ 
\textbf{Refine} & \textbf{0.6389} & \textbf{0.6938} & \textbf{0.7194} &\textbf{0.7451} & \textbf{0.6993} & 16.36 \\ \hline
\end{tabular}    
\label{tab:summarization_strategies}
\end{table*}  

We compare two common summarization strategies for handling multiple retrieved segments: Refine (iteratively updating a single summary, Figure \ref{fig:framework}) and Map-Reduce (parallel segment summarization followed by merging, Figure \ref{fig:map-reduce}).
As shown in Table \ref{tab:summarization_strategies}, Refine consistently achieves higher accuracy across all RETA levels. However, Map-Reduce offers faster processing due to its parallel nature. Therefore, the choice between these strategies depends on the specific application's requirements, prioritizing either accuracy or efficiency.

\begin{table}[ht]  
  \centering  
  \caption{Accuracy comparison for different prompts aimed at enhancing numerical precision.}  
  \fontsize{9}{11}\selectfont  
  \setlength{\tabcolsep}{3pt}  
  \begin{tabular}{l|cccc|c}  
    \hline  
    & \multicolumn{4}{c|}{\textbf{RETA}} & \\  
    & \textbf{0\%} & \textbf{0.001\%} & \textbf{0.01\%} & \textbf{0.1\%} & \textbf{Average} \\  
    \hline  
    \textbf{TD-O} & 0.4917 & 0.4937 & 0.5187 & 0.5750 & 0.5198 \\  
    \textbf{TD-R} & 0.3479 & 0.3479 & 0.3597 & 0.4083 & 0.3660 \\  
    \textbf{TD-S} & 0.4111 & 0.4153 & 0.4493 & 0.5438 & 0.4549 \\  
    \textbf{TD-RS} & 0.4403 & 0.4438 & 0.4722 & 0.5396 & 0.4740 \\  
    \textbf{TD-SP} & 0.5278 & 0.5299 & 0.5479 & 0.5882 & 0.5484 \\  
    \textbf{TD-RSP} & \textbf{0.5646} & \textbf{0.5660} & \textbf{0.5750} & \textbf{0.5938} & \textbf{0.5748} \\  
    \hline  
  \end{tabular}  
  \label{tab:precision_comparison}  
\end{table}  

To improve LLM extraction of precise numerical values, we designed and evaluated six prompt variations (TD-O to TD-RSP), incorporating precision requirements and input-output examples.
TD-RSP, combining precision requirements and a precision-inclusive example, consistently achieved the highest accuracy across all fine-grained RETA levels (Table \ref{tab:precision_comparison}). Conversely, poorly designed prompts (TD-R, TD-S, TD-RS) negatively impacted accuracy compared to a baseline prompt (TD-O), highlighting the importance of careful prompt engineering for numerical precision.

\begin{table}[ht]     
\centering
\caption{Accuracy comparison for different keyword completion settings across various RETA levels.}
\setlength{\tabcolsep}{3pt} 
\fontsize{8}{10}\selectfont 
\setlength{\tabcolsep}{1pt}
\begin{tabular}{l|cccc|c}      
\hline      
& \textbf{RETA 1\%} & \textbf{RETA 3\%} & \textbf{RETA 5\%} & \textbf{RETA 10\%} & \textbf{Average} \\ \hline      
\textbf{K} & 0.3403 & 0.3917 & 0.4076 & 0.4292 & 0.3922 \\ 
\textbf{K\_C} & 0.4681 & 0.5167 & 0.5361 & 0.5604 & 0.5203 \\ 
\textbf{K\_T} & 0.4785 & 0.5396 & 0.5500 & 0.5736 & 0.5354 \\ 
\textbf{K\_T\_C} & \textbf{0.6389} & \textbf{0.6938} & \textbf{0.7194} & \textbf{0.7451} & \textbf{0.6993} \\ \hline      
\end{tabular}      
\label{tab:keyword_completion_comparison}
\end{table}

We investigated the impact of keyword completion on LLM performance by evaluating four settings: \textbf{K} (keyword only), \textbf{K\_C} (keyword + company), \textbf{K\_T} (keyword + time), and \textbf{K\_T\_C} (keyword + time + company).
Table \ref{tab:keyword_completion_comparison} clearly shows that providing additional context through company and time information significantly improves accuracy. K\_T\_C, leveraging the full context, achieves the highest accuracy across all RETA levels, emphasizing the importance of comprehensive keyword completion for effective information extraction.

\begin{table}[ht]  
\centering  
\caption{Accuracy comparison for different numbers of shots across various RETA levels.}  
\fontsize{8}{10}\selectfont  
\setlength{\tabcolsep}{2pt} 
\begin{tabular}{l|cccc|c}
\hline  
& \textbf{RETA 1\%} & \textbf{RETA 3\%} & \textbf{RETA 5\%} &\textbf{RETA 10\%} & \textbf{Average} \\ \hline  
\textbf{0-shot} & 0.4799 & 0.5229 & 0.5354 & 0.5472 & 0.5214 \\ 
\textbf{1-shot} & \textbf{0.6389} & \textbf{0.6938} & \textbf{0.7194} & \textbf{0.7451} & \textbf{0.6993} \\ 
\textbf{2-shot} & 0.6227 & 0.6803 & 0.6966 & 0.7231 & 0.6807 \\  
\textbf{3-shot} & 0.6181 & 0.6806 & 0.7007 & 0.7174 & 0.6792 \\ \hline  
\end{tabular}  
\label{tab:number_of_shots_comparison}  
\end{table} 

Few-shot learning is an important ability of LLMs. To investigate the impact of the number of shots on AIE's performance, we experimented with different numbers of shots, ranging from 0 to 3. 
As shown in Table \ref{tab:number_of_shots_comparison}, the 1-shot setting achieves the highest accuracy across all RETA levels. The performance of 2-shot and 3-shot settings is slightly lower than that of the 1-shot setting but still better than the 0-shot setting. This indicates that a single well-designed example can effectively guide LLMs to generate more accurate responses. However, the slight decrease in performance with additional examples could be attributed to the increased complexity of the input or potential inconsistencies among multiple examples, which may confuse the model rather than provide more guidance.

Based on this experiment, we recommend carefully determining the number of shots when using LLMs for information extraction.
Although providing more shots may still be helpful, it is essential to ensure their consistency and relevance to avoid potential confusion and maintain optimal performance.

\section{Related Work}

Early Information Extraction (IE) relied heavily on rule-based approaches, often targeting specific domains or focusing solely on tables \cite{brito2019hybrid}, thus missing crucial contextual information. Recent machine learning advancements introduced models like bidirectional RNNs for table understanding and BERT for text processing \cite{ma2020spot,chen2019exploring,hillebrand2022kpi}. However, QA-focused approaches like FinQA \cite{chen2022finqa}, TAT-QA \cite{zhu2021tatqa}, and MULTIHIERTT \cite{zhao-etal-2022-multihiertt} only analyze specific sections within HLDs.

This research leverages LLMs for HLD information extraction, addressing three key areas: 1) Long document processing, circumventing input length limitations \cite{liang2023unleashing}. 2) IE \cite{li2023evaluating,wei2023zero}, specifically value extraction, building upon LLM success in NER \cite{gupta2021context,wang2023gpt}, RE \cite{wan2023gpt,xu2023unleash}, and Knowledge Graph Extraction \cite{shi2023chatgraph,polak2023flexible,arora2023language}. 3) Tabular reasoning, utilizing LLMs' prowess in handling structured data \cite{chen2023large,ye2023large}.

\section{Conclusion}

To enable information extraction from Hybrid Long Documents (HLDs) using LLMs, we propose the \textbf{A}utomated \textbf{I}nformation \textbf{E}xtraction (\textbf{AIE}) framework. AIE comprises four modules: Segmentation, Retrieval, Summarization, and Extraction. We introduce \textbf{Fi}nancial Reports \textbf{N}umerical \textbf{E}xtraction (\textbf{FINE}), a dataset constructed from financial reports, to analyze AIE's effectiveness. Extensive experiments on FINE demonstrate the impact of each module and showcase AIE's superior performance compared to baseline methods. Furthermore, we validate AIE's strong performance across diverse domains, including scientific papers and Wikipedia, confirming its generalizability and effectiveness in HLD information extraction.

\bibliographystyle{IEEEtran}
\bibliography{refs}

\end{document}